\documentclass[12pt]{article}

\usepackage{scicite}
\usepackage{times}
\usepackage{siunitx}
\usepackage{graphicx}
\usepackage{wrapfig}
\usepackage{breqn}
\usepackage{float}

\newenvironment{sciabstract}{%
\begin{quote} \bf}
{\end{quote}}

\title{Harnessing bistability for directional propulsion of untethered, soft robots}

\author
{Tian Chen,$^{1\ast}$ Osama R. Bilal,$^{2,3\ast}$ Kristina Shea,$^{1\dagger}$ Chiara Daraio$^{3\dagger}$\\
\\
\normalsize{$^{1}$Department of Mechanical and Process Engineering, ETH Zurich}
\normalsize{8092 Zurich, Switzerland}\\
\normalsize{$^{2}$Institute of Theoretical Physics, ETH Zurich}
\normalsize{8093 Zurich, Switzerland}\\
\normalsize{$^{3}$Division of Engineering and Applied Science, California Institute of Technology}\\
\normalsize{Pasadena, CA 91125, USA}\\
\\
\normalsize{$^\ast$T.C. and O.R.B. contributed equally to this work.}\\
\normalsize{$^\dagger$To whom correspondence should be addressed; E-mail:  daraio@caltech.edu, kshea@ethz.ch}
}

\begin{document}
\baselineskip24pt
\maketitle 

\begin{sciabstract} In most macro-scale robotics systems , propulsion and controls are enabled through a physical tether or complex on-board electronics and batteries. A tether simplifies the design process but limits the range of motion of the robot, while on-board controls and power supplies are heavy and complicate the design process. Here we present a simple design principle for an untethered, entirely soft, swimming robot with the ability to achieve preprogrammed, directional propulsion without a battery or on-board electronics. Locomotion is achieved by employing actuators that harness the large displacements of bistable elements, triggered by surrounding temperature changes. Powered by shape memory polymer (SMP) muscles, the bistable elements in turn actuates the robot's fins. Our robots are fabricated entirely using a commercially available 3D printer with no post-processing. As a proof-of-concept, we demonstrate the ability to program a vessel, which can autonomously deliver a cargo and navigate back to the deployment point. 
\end{sciabstract}

Soft robotics~\cite{Kim2013, Rus2015, Wang2015, Wehner2016, Mazzolai2016} and robotic materials (a.k.a programmable matter)~\cite{McEvoy2015, Goldstein2005, Bilal2017a} are blurring the boundary between materials and machines while promising a better, simpler, safer and more adaptive interface with humans~\cite{Zheludev2012, Reis2015}. Propulsion and navigation are core to both soft and rigid robotic systems. Autonomous (or pre-programmed) propulsion is a central element in the road map for future autonomous systems, enabling, for example,  unguided traversal of open waters (e.g., studying marine biology~\cite{Fernandes2000}, ocean dynamics~\cite{Jaffe2017}). Power supply to enable propulsion remains one of the major obstacles in all forms of locomotion. One of the easiest solutions for supplying power to soft robots is the use of a tether~\cite{Rus2015}. Tethered pneumatics, for example, enabled active agonistic and antagonistic motion~\cite{Shepherd2011} and an undulating serpentine~\cite{Onal2013}. Dielectric elastomers were used to create tethered soft crawlers~\cite{Henke2012} and to simulate the up and down motion of a jellyfish~\cite{Godaba2016}. Electro-magnetics were used to create a tethered Earthworm-like robot~\cite{Song2016}, and untethered microswimmers under a rotating magnetic field~\cite{Tottori2012}. A pressure deforming elastomer was utilized to design an artificial fish tail that can perform maneuvers~\cite{Marchese2014}. Untethered robots, on the contrary, sacrifice simplicity in design for moving freedom without restriction. An untethered robot needs to encapsulate programming, sensing, actuation, and more importantly, an on-board power source.

The demonstration of an entirely soft (composed of materials with elasticity moduli on the order of $10^{4}- 10^9$ Pascal) untethered robot, ``the Octobot''~\cite{Wehner2016}, opened the door to a new generation of robots~\cite{Mazzolai2016}. The Octobot is powered through regulated pressure generating a chemical reaction. Fabrication of the Octobot requires a combination of lithography, moulding, and 3D printing. However, it does not exhibit locomotion. A common feature of all current demonstrations of soft robots is the presence of a complex internal architecture as a result of multi-step fabrication and assembly process. Here, we present a methodology for designing an untethered, entirely soft robot, which can propel itself and can be pre-programmed to follow selected trajectories. Furthermore the robot can be preprogrammed to reach a destination, deliver a cargo and then reverse its propulsion direction to return back to its initial deployment point. The robot can be fabricated using a commercially available 3D printer in a single print with no assembly. However, it is partitioned to study the behavior of the different components.

Within the scope of soft robotics~\cite{Rus2015}, we focus on the actuation, design and fabrication of a robot that exploits bistable actuation for propulsion, and responds to temperature changes in the environment, to control its directional locomotion. We use shape memory polymers (SMPs) to create bistable ``muscles'' that respond to temperature changes in the environment. Bistable actuation is often found in biological systems, like the Venus fly-trap~\cite{Skotheim2005} and the Mantis shrimp~\cite{patek2004biomechanics}. When working near instabilities, bistability can amplify displacements with the application of a small, incremental force~\cite{Raney2016}. Engineers started to integrate instabilities in design~\cite{Reis2015}, for example, in space structures~\cite{Schioler2007}, energy absorption mechanism~\cite{Restrepo2015}~\cite{Shan2015} and fly-trapping robots~\cite{Zhang2016}. By amplifying the response of soft SMP muscles, snap-through instabilities can instantaneously exert high force and trigger large geometrical changes~\cite{Overvelde2015}. Bistability has also been utilized to sustain a propagating solitary wave in a soft medium~\cite{Raney2016}. More recently bistability enabled the realization of the first purely acoustic transistor and mechanical calculator~\cite{Bilal2017}. A typical bistability is found in the Von Mises truss design, which allows a simple 1D system to have two stable states~\cite{Mises1923}. Combining this principle with multi-material 3D printing, it possible to realize monolithic, bistable actuators with a tunable activation force through material and geometry changes~\cite{Chen2017}. These actuators can be used to create load-bearing, multi-state reconfigurable 3D printed structures, where large shape changes are possible due to the long stroke length of the bistable actuator design. To autonomously activate them, they can be combined with 3D printed shape memory strips, which respond to different temperatures, to create time sequenced linear actuators and the first 4D printed, deployable structures~\cite{Chen2017a}. Here, we build on these works to create untethered robots.

\begin{figure}
	\begin{center}
		\includegraphics[width=.65\textwidth]{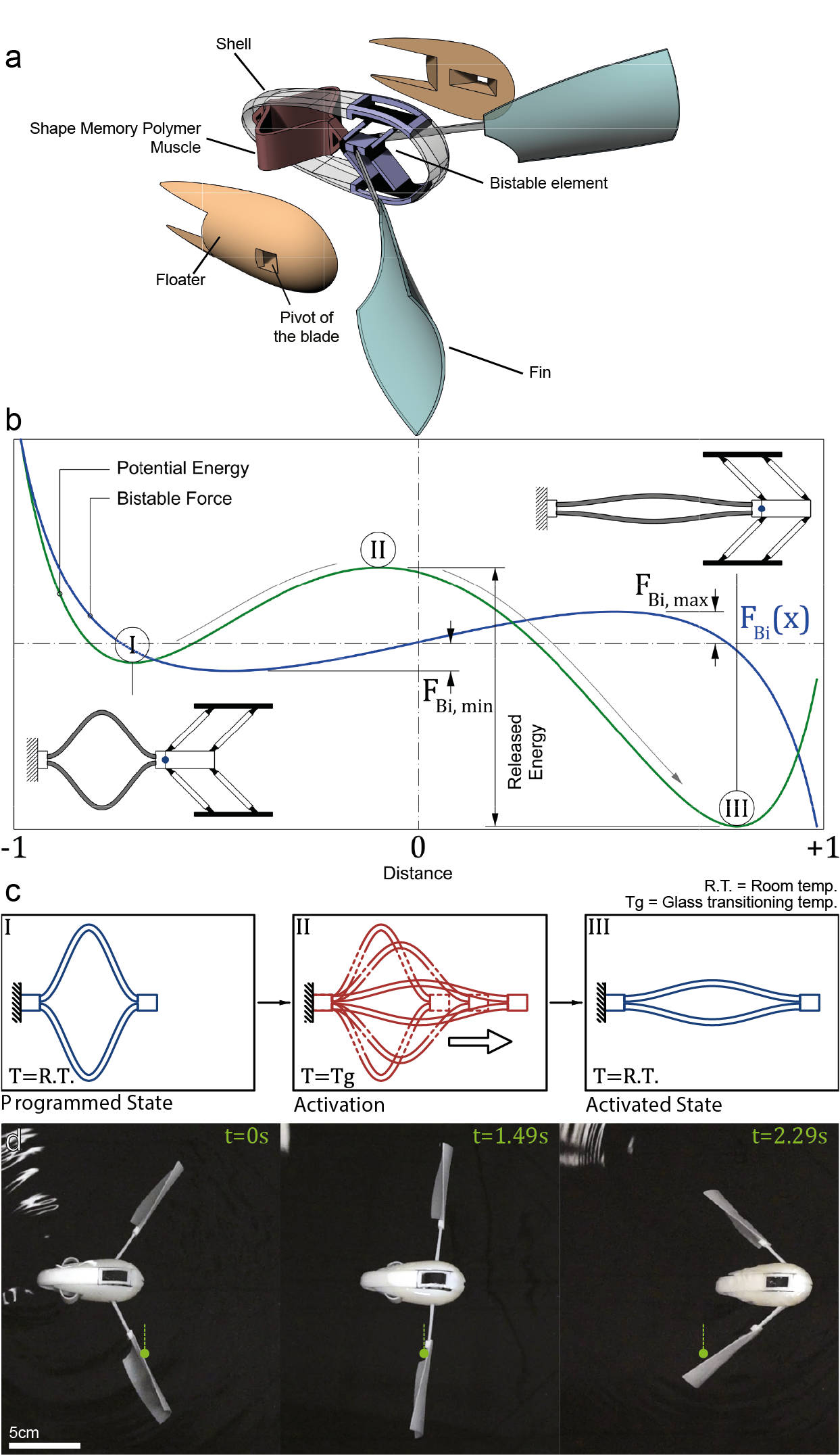}
	\end{center}
	\caption{Propulsion through bistability: (a) schematic of a 3D printed, soft robot (parts are false colored for visualization) (b) Energy potential of the bistable element with two stable states, I and III. The asymmetry in the curve indicates the need for larger amount of energy to move backward than forward. The SMP ``muscle'' shown in the inset are rotated \SI{90}{\degree} with respect to the bistable element for visualization. (c) The SMP muscle in the deployed (I), transitioning (II) and activated phases (III). (d) Screen captures of the deployed robot in temperature ($T\ge T_{\mathrm{g}}$) at the different phases of activation.}
	\label{fig:1}
\end{figure}

\section*{The robot design principle}


To demonstrate the underlying principle of our design concept, we first study a robot propelled by a single actuator~(Fig. \ref{fig:1}). The actuator consists of two strips of SMP material, acting like a ``muscle'' connected to a bistable element. To reduce design complexity and ease prototyping, the robot is decomposed into five parts (outer shell, floaters, fins, bistable element and shape memory ``muscle'' (Fig. \ref{fig:1}a)). The shell supports the bistable element to ensure linear actuation and provide stability for the robot. The floaters insure that, in the vertical direction, the SMP strips are fully submerged in water. The groove in the floaters provides the pivot point for the fins. The fins are attached to the bistable element with elastomeric joints, ensuring flexibility.

The propulsion mechanism follows the forward motion of many organisms with fins submerged in water~\cite{Song2016a}. The fins, which perform a paddling motion, are driven primarily by destabilizing buckling trusses. The SMP muscle utilized to trigger the bistable element is a pair of 3D-printed curved beams that exhibit shape memory behavior~\cite{Wagner2017}. One limitation of shape memory materials is their slow activation speed, therefore we attach the muscle to a bistable element (Fig. \ref{fig:1}b) with the fins mounted on it. In addition to amplifying the output force~\cite{Chen2017a}, the bistability transforms the slow, small motion of the SMP beams into a rapid, amplified one propelling the robot forward. It is often recognized that in a physical implementation of a bistable element using compliant joints, the two equilibrium states are not equally energetic, i.e., $\vert F_{\mathrm{Bi, min}}\vert<\vert F_{\mathrm{Bi, max}}\vert$. Due to the rotational stiffness of the flexible joints, the fabricated state is more stable than the activated state~\cite{Chen2017}. While this is a disadvantage when constructing multi-stable systems, it can be exploited when directional transformation is desirable~\cite{Raney2016}. 

Before deploying the robot, we heat the printed SMP muscle past its glass transition temperature ($T_{\mathrm{g}}$) and mechanically deform it to the programmed shape. After deploying the robot in water (state I) with temperature equal-to or larger-than ($T_{\mathrm{g}}$), the muscle relaxes, transforming back into its original/printed shape (state II). Since this is a constrained relaxation, the muscle must overcome the activation force of the bistable element at all points between state I and II, i.e. $F_{\mathrm{SMP}}(x)>F_{\mathrm{Bi}}(x)~~\forall x = -1, \dots, 0$. Post actuation, the system transitions into state III.

The vertical equilibrium of the robot is achieved by balancing buoyancy and weight forces, in-plane acceleration occurs when propulsion overcomes drag. The forward motion of the robot occurs in the transition between state I and III, consisting of prior- and post-snapping (Fig. \ref{fig:1}b). Prior to the onset of the instability (Fig. \ref{fig:1}d), the shape memory muscle moves the bistable element and drives the fins backward until they are perpendicular to the direction of motion (state II). This is a relatively slow motion with low energy output. Immediately after, the bistable element snaps to its second equilibrium position. This drives the fins rapidly and increases the velocity of the robot. 

\section*{The bistable element-muscle pair}

\begin{figure}
		\begin{center}
	\includegraphics[width=1\textwidth]{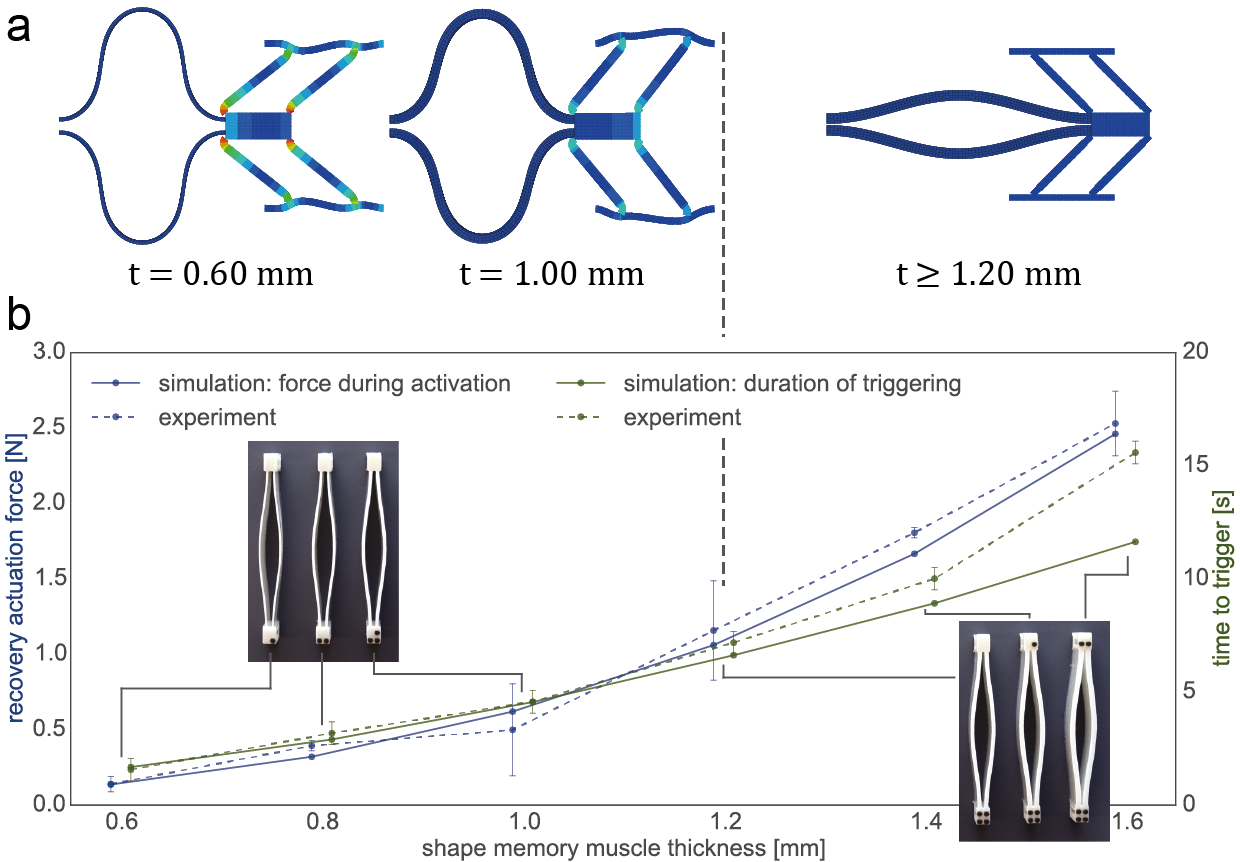}
\end{center}
	\caption{Actuator design: (a) FEM simulations of the constrained recovery of the bistable-muscle pair. The vertical dashed black line separates thicknesses unable to activate the bistable element (left) from functional thicknesses (right). (b) Experimental and numerical correlation between the thickness of the SMP muscle and its recovery forces as well as the time it takes to heat to its original shape. The inset shows the different muscles tested. The error bars in the force readings represent the standard divination. The error bars for the activation times represent the error in reading the times from video recordings of the experiments.}
	\label{fig:2}
\end{figure}

To trigger the instability, the actuation force must overcome the energy barrier of the bistable element (Fig. \ref{fig:1}b). In order to determine the range of operational actuation forces, we use the finite element method to simulate the constrained recovery of the actuation pair (i.e., the bistable element and the SMP muscle). We vary the thickness of the SMP beams between \num{0.6} and \SI{1.6}{\milli\metre}. Full recovery (i.e., snapping of the bistable element) does not occur for beams with thicknesses lower than \num{1.2} mm (Fig. \ref{fig:2}a). Thinner beams induce localized stresses in the bistable element, although not enough to cause it to snap. The recovery forces of the muscle range from \num{0.2} to \SI{2.1}{\newton} in simulation (Fig. \ref{fig:2}b). However, we experimentally observe slightly larger forces at higher thickness values due to the fusion of the two SMP beam strips at both ends of the muscle during polymerization. To determine the activation time needed, we simulate the actuation process using an invariant boundary temperature, while calculating the time for the muscles to reach thermal equilibrium through conductive heating. To characterize the activation time experimentally, six SMP muscles were first programmed and then submerged in hot water simultaneously. The time of activation is extracted from the recorded video (See SI Movie 7). At equal length, the force supplied by the muscles as well as the actuation time increases with thickness (Fig. \ref{fig:2}b). 

It is worth noting, however, that the distance traveled by the robot depends mainly on the bistable element rather than the muscle. This is demonstrated by fabricating and testing the same robot with different muscle thicknesses (SI Fig. 1). Regardless of the force output of the muscle the robot travels the same distance (\SI{115}{\percent} of its length). Such demonstration shows the significance of the bistability and long stroke length for the propulsion of soft robots, regardless of the initial actuation force. Since inducing a stroke depends on overcoming the bistable energy barrier rather than the actuator force, the bistable element can be designed with a very small energy barrier~\cite{Bilal2017} and high asymmetry~\cite{Raney2016}. This is also promising for the miniaturization of the actuators and the increase of the number of actuators in a given robot.

\section*{Sequential and directional propulsion}

\begin{figure}
		\begin{center}
	\includegraphics[width=.8\textwidth]{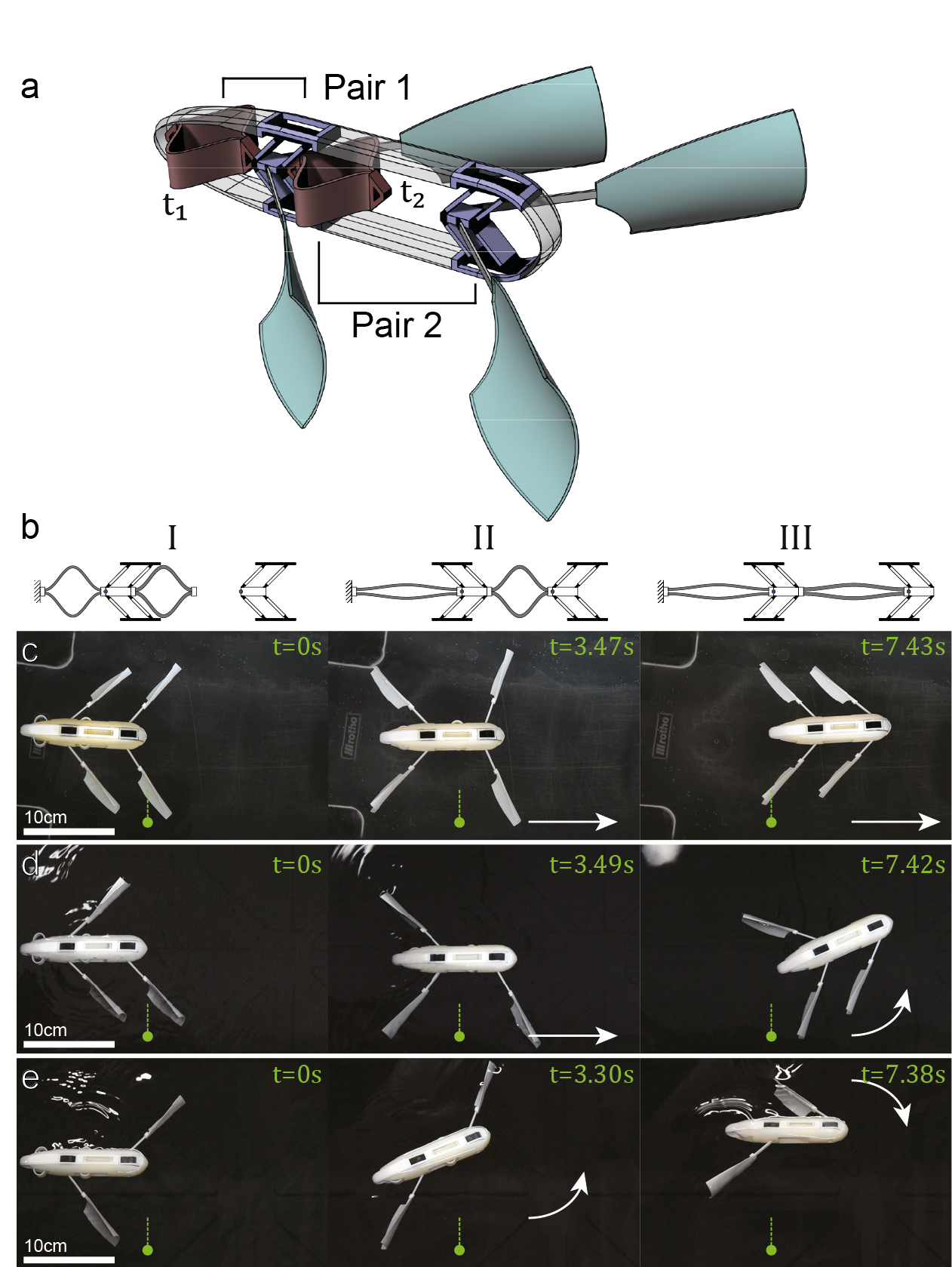}
		\end{center}
	\caption{Sequential and directional propulsion: (a) A schematic of a robot with two bistable element-muscle pairs. A thinner muscle with faster actuation time is placed at the rear, i.e., $t_1<t_2$. b. Three distinct phases of the actuation. (b) Activation sequence: (I) initial state with both muscles programmed. (II) the thinner muscle activates, triggering the first bistable element and pushing the second muscle to touch the second bistable element. (III) The second (thicker) muscle triggers the second bistable element. (c-e) Snapshots from deploying three configurations of the multi-stroke swimmer showing three different directional motion at each of the three phases.}
	\label{fig:3}
\end{figure}

The use of bistability for propulsion can be expanded from a single-stroke to a multi-stroke robot (Fig. \ref{fig:3}a). Multiple bistable element-muscle pairs can be compacted into the same robot in either a connected or separate fashion. If $N$ pairs are completely disconnected, they can induce $N$ independent strokes that are either simultaneous (using identical pairs) or sequenced (using different ones). If the pairs are connected in series (e.g., the muscle of pair two is attached to the bistable element of pair one), they can create synchronized strokes based on the propagation of a soliton-like actuation~\cite{nadkarni2016unidirectional} (Fig. \ref{fig:3}b). This guarantees that a stroke $n$ will always be executed before the $n+1$ stroke. 

Post-activation of actuation pair one, muscle two is moved into position to activate the bistable element of pair two (Fig.~\ref{fig:3}b, II). As the combined force of the already-activated bistable element and the muscle from pair one is greater than the actuation force of muscle two, i.e., $F_{\mathrm{Act},1}+F_{\mathrm{Bi, max}}>F_{\mathrm{Act},2}>F_{\mathrm{Bi, min}}$, muscle two displaces more in the forward direction and activates the second bistable element. This creates a forward chain action. 

In addition to the increased net forward motion, due to the added actuation pairs, propulsion in various directions can be achieved by adjusting the placement of the fins. Having two fins (one on each side) induces a symmetric moment that moves the robot forward. When one of the fins is removed, an asymmetric moment arises, giving the robot a push towards the direction of the missing fin. By strategically placing the fins, navigation can be preprogrammed in advance, setting the robot on a predefined path. The sequential nature of actuation allows for predictable directional change at each stage. 

To demonstrate the sequential forward motion of the robot, four fins are placed symmetrically on the sides of two actuation pairs. The force needed to trigger the bistable element is identical to that of the single stroke robot. The thickness of the rear muscle is \SI{1.2}{\milli\metre}, while the front one is \SI{1.6}{\milli\metre}, to have a sufficient time gap between the two activations. Once deployed, two consecutive forward motions are executed, increasing the overall distance traveled to \SI{190}{\percent} of a single stroke robot length (Fig. \ref{fig:3}c). Next, we remove the front left fin and deploy the robot again (Fig. \ref{fig:3}d). The rear (symmetric) actuation induces a forward motion followed by a left turn with $\approx$ \SI{23.85}{\degree}. Afterwards, we deploy the robot with only two fins placed asymmetrically in the front-left and rear-right position (Fig. \ref{fig:3}d). The robot takes a left turn of $\approx$ \SI{21.64}{\degree} followed by a right one of $\approx$ \SI{-21.45}{\degree}. By designing different sizes of fins, one can control the turn angle of the robot.


\begin{figure}
	\includegraphics[width=1\textwidth]{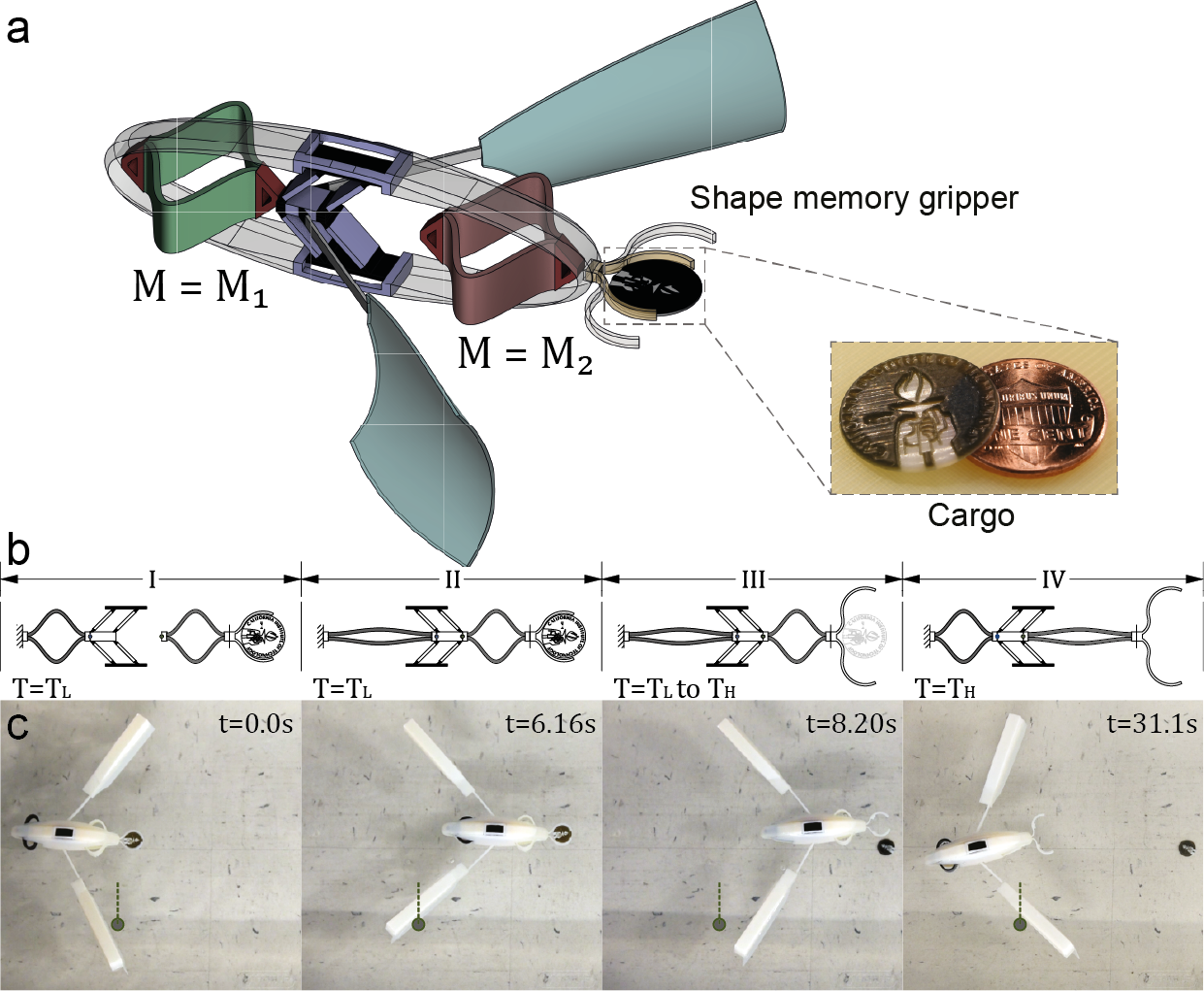}
	\caption{Reverse navigation: A schematic of a robot with two muscles and a single bistable element. The first muscle, $M_1$, is fabricated with a material that activates at $T_{\mathrm{L}}$. The second muscle, $M_2$, activates at $T_{\mathrm{H}}$, which is higher than $T_{\mathrm{L}}$. (b) Sequence of activation, (I) at room temperature, $T_{\mathrm{R.T.}}$, both muscles are programmed. (II) As water temperature increases to $T_{\mathrm{L}}$, the first muscle triggers the bistable element and propels the swimmer forward. (III) As water heats to $T_{\mathrm{H}}$, the grippers relax and release the cargo. (iv) When the water temperature reaches $T_{\mathrm{H}}$, the second muscle reverses both the bistable element and re-programs the first muscle. (c) Snapshots from the deployment of the robot demonstrating the four different phases.}
	\label{fig:4}
\end{figure}

\subsection*{Cargo Deployment and Reverse Navigation} 
In many scenarios, such as cargo deployment or retrieval, a robot is required to navigate back to the starting point of its journey after the required operation is completed. The sequential and directional motion of the robot presented so far is based on identical muscles (i.e., same material) with varying thickness. This thickness variation translates into different activation times at the same temperature. 

In order to design our robot to reverse its navigation path, we incorporate an extra muscle on the other side of the bistable element (Fig. \ref{fig:4}a-b). For the second set of muscles (i.e., for reverse navigation) we add an extra dimension to the design, that is the activation temperatures. As it has been shown that by varying the constituting inkjet materials, one can achieve different glass transition temperatures~\cite{Mao2015}. Utilizing this property, we fabricate two identical muscles with two different materials and therefore activation temperatures $T_{\mathrm{L}}$ and $T_{\mathrm{H}}$. The second muscle induces enough force to overcome both the bistable potential and the first muscle. Such force can trigger the instability producing a force in the opposite direction to the initial actuation, causing the robot to reverse its navigation direction.

To demonstrate cargo delivery and robot return, we add a shape memory gripper at the front of the robot to hold a 3D-printed penny (Fig. \ref{fig:4}a-b). Once the robot is deployed in water with temperature, $T_{\mathrm{L}}$, the first muscle activates and triggers the bistable element, thus propelling the swimmer forward to the delivery point (Fig. \ref{fig:4}c). Once the water temperature reaches the glass transition temperature of the gripper, it relaxes to its original shape and releases the penny. When the temperature reaches $T_{\mathrm{H}}$, the second muscle triggers the bistable element in the reverse direction causing the robot to move backwards to its original deployment point. The principle of reverse propulsion can be easily  extended to more complex trajectories and multiple cargoes.
 
\section*{Outlook}
This work presents an entirely soft, swimming robot that requires neither complex on-board components nor a tether to achieve preprogrammed directional propulsion. Instead, programmed shape memory polymer muscles are used as the power supply, a bistable element as the ``engine'', and a set of fins as the propellers. By reacting to varying external temperatures through design of the geometry and material of both the bistable element and shape memory strip, the robot can move forward, turn, reverse and/or deliver a cargo. This demonstrates a first step in the realization of entirely-soft locomotive robots, potentially applicable in a variety of applications such as navigation and delivery.

\section*{Materials and Methods}
All components are fabricated with the multi-material Stratasys Connex printers. The SMP muscles are fabricated with VeroWhitePlus plastic ($T_{\mathrm{g}}\simeq\SI{60}{\degree}$), with the exception of $M_1$ in Fig.~\ref{fig:4}, which is fabricated with FLX9895 ($T_{\mathrm{g}}\simeq\SI{35}{\degree}$). The compliant components within the bistable element are fabricated with Agilus30 with ($T_{\mathrm{g}}<\SI{-5}{\degree}$) and therefore remains in the rubbery state throughout the experiments. All remaining components are fabricated with a high temperature resistant material, RGD525 ($T_{\mathrm{g}}>\SI{80}{\degree}$), which retains its stiff throughout the activation process~\cite{Wagner2017}. The stiffness of the materials used in the robot are in the range of $2 \times 10^{6}- 10^9$ Pascal.
The glass transition temperature of FLX9895 is approximately \SI{35}{\degree} and \SI{60}{\degree} for VeroWhitePlus. Finite element simulations are implemented using Abaqus 14.1, geometrical non-linear solver with thermal-viscoelastic material model.

\section*{Acknowledgments}
The authors would like to thank Jung-Chew Tse for fabrication support, Connor McMahan and Ethan Pickering for support in experiments. This work was partially supported by the Army Research Office, Grant number W911NF-17-1-0147 and by an ETH Postdoctoral Fellowship FEL-26 15-2 (to O.R.B.).

\bibliographystyle{Science}
\bibliography{swimmer-manuscript}

\section*{Appendix}

\subsection*{Bistable Mechanism Analysis}
\begin{figure*}
	\includegraphics[width=\textwidth]{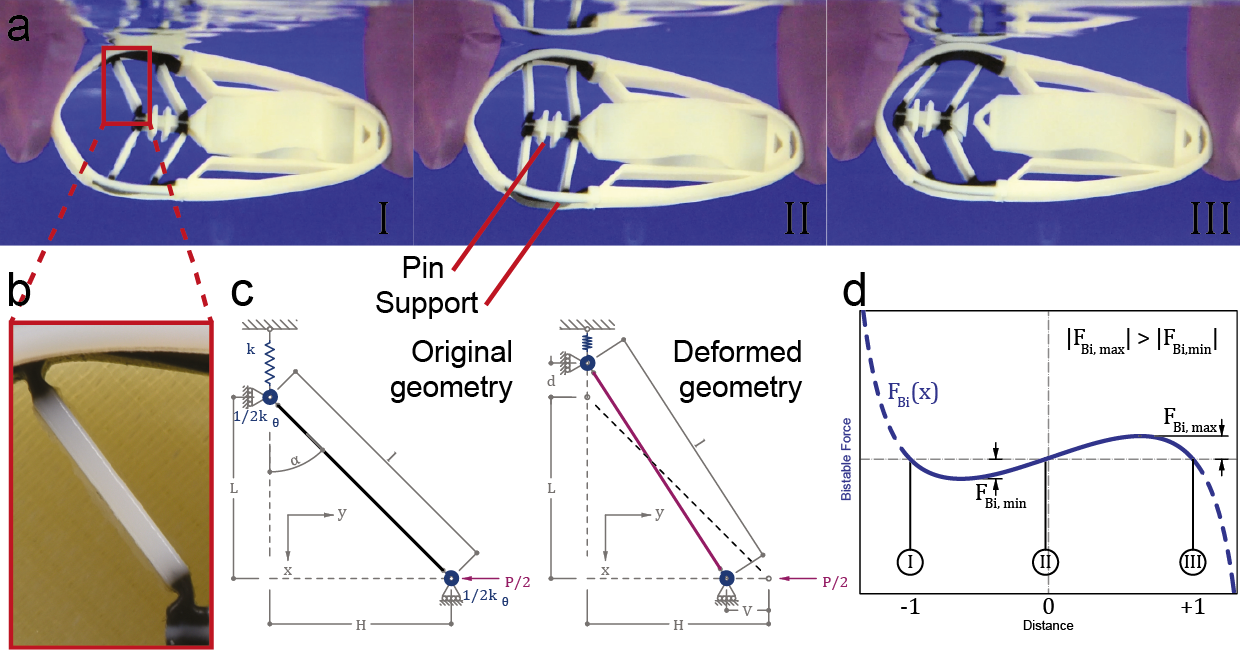}
	\caption{  a. Video snapshot of the actuation of the bistable mechanism. Note that from stage II to III, the shape memory polymer does not contribute to propulsion. b. A zoom in view of the bistable truss geometry, the black regions represent the flexible material, and the white is the rigid material. c. An idealization of the truss geometry in both the original and the deformed state. The support is idealized by a linear spring $k$, and the joints by a torsional spring $k_\theta$. d. The force displacement relationship of the bistable mechanism as derived in equation~\ref{eq:6}.}
	\label{fig:SI_1}
\end{figure*}

In order to analyze the bistable mechanism, we first fabricate a bistable mechanism-muscle pair enclosed in a planar shell identical to that of a single stroke swimming robot (Supp. Fig.~\ref{fig:SI_1}.a). The four bars within the bistable mechanism form two ``V" shapes pointing to the right, representing one of the stable states of the mechanism. The sample (the pair confined in the planner shell) is then immersed in \SI{60}{\degreeCelsius} water (state I), equivalent to the glass transition temperature of the muscle material (VeroWhitePlus). As the muscle temperature rises due to water contact, both of the deformed beams within the muscle start relaxing into their original/printed shape. At the onset of the instability, where the truss-like bars of the bistable mechanism are vertical (state II), the muscle pushes the mechanism slightly towards its second stable state (where the ``V" shapes are pointing to the left) (state III). 

In order to model the bistable behavior of the mechanism, we now consider one of its four bars (Supp. Fig.~\ref{fig:SI_1}.b). The bars are printed with VeroWhitePlus with Young's modulus E = $2 \times 10^{9}$ Pa, while its connections to the planar shell are printed with Aguils300 with a Young's modulus E = $2 \times 10^{6}$ Pa. Since the bar material has a stiffness that is $10^{3}$ higher than its connection points, we model the bar as an inclined rigid truss element (Supp. Fig.~\ref{fig:SI_1}.c) supported by two torsional springs with a spring constant ($k_{\theta}$) and a linear spring with a constant ($k$) (Supp. Fig.~\ref{fig:SI_1}.c). The rotational springs $k_\theta$ are the torsional resistance of the compliant joints. The linear spring, $k$ simulates bending of the flexible support through linear force in the $y$ direction. 

We use a Lagrangian equation to construct a relationship between the force $P$ and the corresponding displacement $V$ with respect to the deformed geometry. The bar is assumed to be axially rigid. 

\begin{equation}
\mathcal{L}=\frac{1}{2}kd^2+\frac{1}{2}k_\theta \Delta\alpha^2-\frac{1}{2}PV
\end{equation}

Then we consider the deformed geometry with a rigid truss bar, to relate the different displacements in the model:

\begin{equation}
\sqrt{H^2+L^2}=\sqrt{(H-V)^2+(L+d)^2}
\end{equation}

The solution of $d$ is the difference between the deformed projected length and the original one,

\begin{equation}
d=\sqrt{2HV+L^2-V^2}-L
\end{equation}

To simplify the representation, we denote the first term of the solution as $$L_1=\sqrt{2HV+L^2-V^2}.$$ The Lagrangian equation becomes

\begin{equation}
\mathcal{L}=\frac{1}{2}k\left(L_1-L\right)^2+\frac{1}{2}k_\theta \left (\arctan{\frac{H-V}{L_1}}-\arctan{\frac{H}{L}} \right)^2-\frac{1}{2}PV
\end{equation}

We obtain the relationship between $P$ and $V$, by differentiating the system w.r.t. to $V$ and setting the result to zero, $\frac{\partial\mathcal{L}}{\partial V}=0$. Such an equation provides the means to assess the impact each variable on the overall behavior of the system has, and therefore design the bistable mechanism.

\begin{dmath}\label{eq:6}
	P=-2\frac {1}{L_1} \left[ k(L-L_1)(H-V)+ k_\theta\left(\arctan \left( {\frac {H-V}{L_1}} \right) -\arctan \left( {\frac {H}{L}} \right)  \right) \right]
\end{dmath}

In the initial state (I), The shape memory muscle doesn't exert any force $P=0$ on the bistable mechanism (Supp. Fig.~\ref{fig:SI_1}.d).  As the surrounding water heats the muscle, it starts to relax to its original/printed shape, pushing the bistable truss towards its second stable state. Until the muscle pushes the truss to be vertical, i.e. $H=0$ (II), beyond this point, the mechanism flips to the second equilibrium state (III), where the muscle is physically detached from the mechanism.


\subsection*{The role of bistability vs muscles in propulsion}

\begin{figure}[h]
	\includegraphics[width=\textwidth]{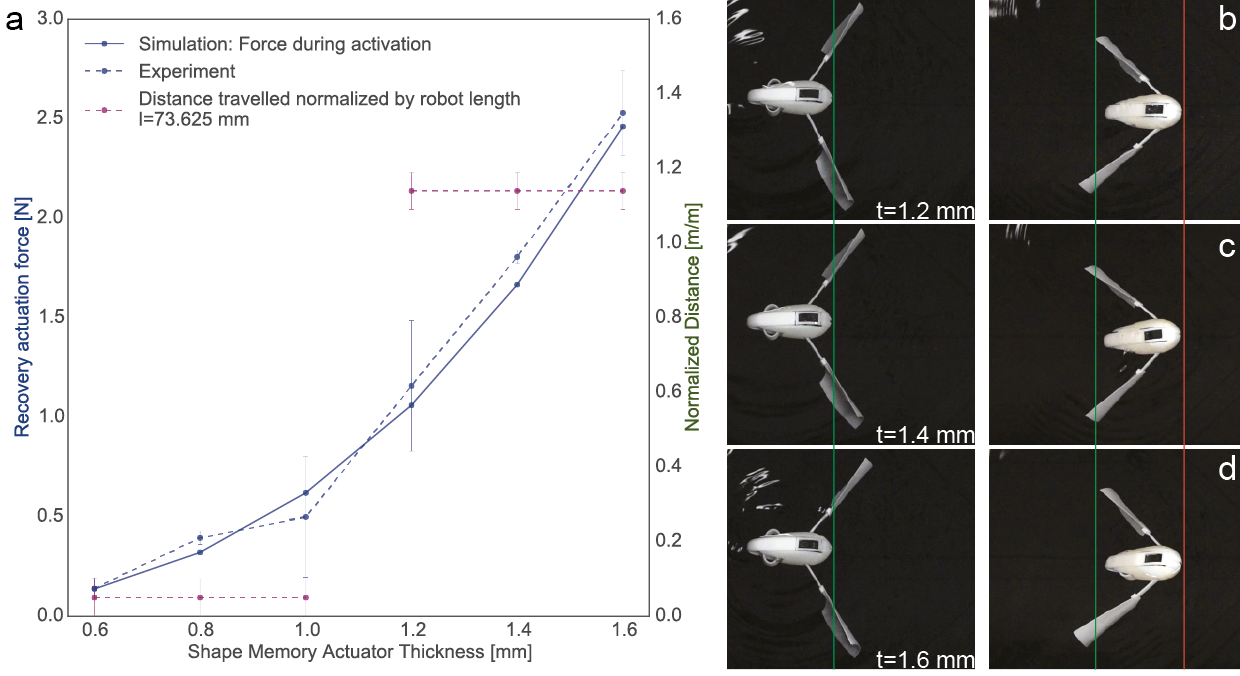}
	\caption{Three swimmers travel the same distance when the shape memory actuator of three different beam thicknesses are used. This shows that propulsion comes predominantly from triggering of the bistable mechanism. The beam thickness values are \num{1.2}, \num{1.4}, \SI{1.6}{\milli\metre}.}
	\label{fig:SI_2}
\end{figure}

In order to assess the contribution of the muscle-induced force on the distance traveled by the swimming robot, we deploy the same robot with various muscles. We systematically increase the thickness of the beams within the muscle, therefore increasing the resultant force  (Supp. Fig.~\ref{fig:SI_2}.a). For beams with thickness $<$ 1.2 mm, the robot did not move forward, as the muscle force is not strong enough to overcome the bistability energy  barrier. All the muscles with beams $>$ 1.2 mm overcame the energy barrier and were able to propel the robot forward. However, the robot traveled the same distance (Supp. Fig.~\ref{fig:SI_2}.b), regardless of the increase of the force amplitude by a factor 2. Therefore, the distance traveled by the swimmer depends on the bistable element rather than the muscle force, as long as the muscle is strong enough to push the mechanism to the onset of the instability.

\pagebreak

\subsection*{Movies}
\begin{description}
	\item[SI Movie 1] Propulsion of a single stroke swimmer. The distance travelled is approximately $1.15l$ where $l$ is the body length of the swimmer.
	\item[SI Movie 2] Propulsion of a two-stroke swimmer. The sequence of activation is controlled by the thickness of the shape memory muscle. The distance travelled is approximately $1.9l$ (where $l$ is the length of the single actuator swimmer).
	\item[SI Movie 3] Propulsion of a two-stroke directional swimmer. The programmed path is straight followed by a left turn. The distance travelled is approximately $0.5l$ after the first stroke, and a turn of $\SI{23.85}{\degree}$ after the second stroke.
	\item[SI Movie 4] Propulsion of a two-stroke directional swimmer. The programmed path includes a left turn followed by a right turn. The rotation is approximately $\SI{21.64}{\degree}$ after the first stroke, and $\SI{-21.45}{\degree}$ after the second stroke.
	\item[SI Movie 5] Propulsion of a reversing swimmer. The programmed operation includes a forward stroke, deployment of cargo, and a reverse stroke. The sequence of activation is controlled by the temperature of the surrounding environment. The first stroke is activated when water reaches $T_{\mathrm{L}}$, deployment occurs when water is heated to $T_{\mathrm{H}}$, then the reverse stroke occurs.
	\item[SI Movie 6] Internal mechanism of the actuator showing the shape memory muscle pushing the bistable mechanism from one equilibrium state to the next.
	\item[SI Movie 7] Shape memory muscle of different thickness exhibiting different time to activation.
	\item[SI Movie 8] Shape memory muscle of different material exhibiting controlled activation depending on the surrounding temperature.
\end{description}
\end{document}